\documentclass{article}

\usepackage[preprint]{neurips_2026}

\usepackage[utf8]{inputenc}
\usepackage[T1]{fontenc}
\usepackage{hyperref}
\usepackage{url}
\usepackage{booktabs}
\usepackage{amsfonts}
\usepackage{amsmath}
\usepackage{graphicx}
\usepackage{float}
\usepackage{algorithm}
\usepackage{algpseudocode}
\usepackage{nicefrac}
\usepackage{microtype}
\usepackage{xcolor}
\usepackage{multirow}
\usepackage{enumitem}
\usepackage{placeins}
\usepackage{wrapfig}
\usepackage{caption}
\usepackage{capt-of}
\usepackage{subcaption}

\title{\textsc{Forced Deferral}: Manipulating Routing Decisions in Multimodal LLM Cascades}

\author{
Zhongye Liu, Yaopei Zeng, Yurui Chang, Lu Lin\\
Pennsylvania State University\\
\texttt{\{zql5622, ypz5549, yuruic, lulin\}@psu.edu}
}

\begin{document}

\maketitle

\begin{abstract}
    While multimodal large language models (MLLMs) have shown strong visual reasoning abilities, serving a large model for every query is computationally expensive. MLLM cascades mitigate this cost by first querying a weak but cheaper model and deferring to a strong model when the weak model's output is unconfident.
    However, since the weak model's confidence directly controls compute allocation, these systems expose a new attack surface: an adversary can manipulate confidence so that their queries are consistently deferred to the strong model.
    Motivated by this vulnerability, we introduce the Forced Deferral Attack (FDA), an adversarial image attack that lowers the weak model's confidence and causes cascades to route queries to the strong model. 
    FDA learns a universal border trigger by optimizing a temperature-flattened objective. This objective pushes the weak model's token distribution on triggered inputs toward less concentrated targets constructed from its clean responses.
    Across datasets, model families, and deferral metrics, FDA consistently increases strong-model routing while outperforming image-perturbation and prompt-injection baselines. These results show that MLLM cascades are vulnerable to attacks that manipulate compute allocation, forcing unintended strong-model usage without directly targeting answer correctness. 
\end{abstract}

\section{Introduction}

Multimodal large language models (MLLMs) have become increasingly capable at answering questions about images, enabling applications such as visual question answering~\citep{goyal2017making}, document understanding~\citep{mathew2021docvqa}, and interactive assistants~\citep{achiam2023gpt,liu2023visual}. This capability comes at substantial inference cost, especially when large MLLMs are deployed for every query.
To balance inference cost and answer quality, recent systems increasingly rely on efficient model selection mechanisms such as routing~\citep{ong2024routellm,ding2024hybrid} and cascading~\citep{chen2023frugalgpt,yue2023large}.
In an MLLM cascade system, a cheaper weak model first handles the query, and a stronger model is invoked only when the weak model's answer appears uncertain. By reserving expensive inference for only difficult or uncertain cases, cascades allow the model provider to improve inference efficiency while preserving answer quality. 
Despite their efficiency benefits, MLLM cascades expose a compute-allocation attack surface. They are deployed to reserve strong-model computation for difficult queries, but a user may prefer strong-model responses even for queries that the provider intends to serve with the weak model. This incentive mismatch motivates an attack on the confidence signal: by lowering the weak model's confidence, an adversary can force unnecessary deferrals to the strong model. The adversary thereby obtains higher-quality responses while shifting the additional compute cost to the provider. Unlike attacks that aim to degrade answer correctness, this threat targets the cascade's allocation mechanism rather than the final prediction.
This vulnerability highlights the gap in existing studies of efficient inference and adversarial robustness. Research on efficient model selection has primarily focused on the cost--quality tradeoff under benign deployment~\citep{ong2024routellm, chen2023frugalgpt}. The robustness of the deferral decision itself has received little attention. Prior attacks~\citep{shafran2025rerouting, lin2025life} on model allocation have mostly considered pre-generation routing, where the model is selected before any answer is produced. In parallel, adversarial studies of MLLMs typically target output-level behavior, such as causing the model to produce incorrect answers~\citep{zhao2023evaluating, schlarmann2023adversarial} or bypassing safety alignment~\citep{qi2024visual}. None of these lines of work captures the robustness of confidence-based MLLM cascades, where the deferral decision itself becomes an attack surface.

This gap motivates our central research question: \textit{Can an adversary manipulate the weak model's confidence to force unnecessary deferral to the strong model?} Answering this question creates two challenges. 

First, the attacker has no prior knowledge about which uncertainty metric the deployed cascade uses, so the attack cannot be tailored to a single confidence score. Second, the attack must lower the weak model's confidence without damaging the query content, since the attacker still wants the strong model to answer correctly.

To address these challenges, we propose a cascade deferral attack that induces strong-model routing by reducing the weak-model confidence while preserving the input semantics. 
 
To handle unknown deferral metrics, the attack constructs less concentrated target distributions from the weak model's clean output trajectory and optimizes the trigger so that the weak model's token distributions on triggered inputs match these flattened targets. This drives confidence down in a metric-agnostic way rather than against any single score. To preserve the input semantics, the trigger is constrained to a universal border region around the image, leaving the central visual content intact.
Across multiple datasets, model families, and deferral metrics, our attack substantially increases strong-model routing and outperforms common image-perturbation and prompt-injection baselines.

To summarize, we make the following contributions:
\begin{itemize}[leftmargin=*]
  \item We identify and formulate the cascade deferral attack as a new efficiency oriented threat to MLLM cascades. Unlike standard adversarial attacks that aim to change model predictions, this attack manipulates the weak model's confidence to increase the deferral rate to the stronger model, undermining the computational benefits of cascaded inference. 
  \item We propose an answer-agnostic universal border trigger for cascade deferral attacks. By using a temperature-flattened teacher-forcing objective, the trigger reduces weak model's confidence along clean output trajectories, enabling metric-agnostic deferral manipulation while preserving the strong model's performance on triggered inputs.  
  \item We empirically demonstrate that the proposed attack is effective, transferable, and robust across datasets, model families, deferral metrics, and preprocessing defenses, consistently increasing strong-model deferral while preserving answer quality on triggered inputs. 
  
\end{itemize}

\section{Related Work}

\paragraph{Efficient Adaptive Inference with Routing and Cascades}

Large MLLMs are expensive to serve, especially when visual queries
vary in difficulty. Efficient adaptive inference reduces this cost by
assigning queries to models with different capability and cost
profiles. Routing methods select a model before generation, often
using the query, model metadata, or a learned
router~\citep{ong2024routellm, ding2024hybrid}. Cascading methods
call models sequentially: a cheaper model answers first, and a
stronger model is invoked only when the cheaper model's answer
appears unreliable~\citep{chen2023frugalgpt, yue2023large, aggarwal2023automix}.
A key component of cascades is the deferral rule, which decides
whether to accept the weak model's answer or defer to a stronger
model.  
Existing works make the deferral decision based on  estimated uncertainty signal on the weak model using token probabilities~\citep{mahaut2024factual,kadavath2022language}, perplexity-based confidence~\citep{fomicheva2020unsupervised},
verbalized confidence~\citep{tian2023just}, or consistency across
sampled outputs~\citep{lin2023generating}.

\paragraph{Adversarial Robustness of Model Allocation and MLLMs}

Adversarial attacks on MLLMs and vision language models often modify
the visual input to change model behavior. Common forms include
imperceptible perturbations~\citep{zhao2023evaluating, schlarmann2023adversarial}, localized patches~\citep{brown2017adversarial},
and universal triggers~\citep{moosavi2017universal}. These attacks
are usually evaluated by their effect on final answers, such as
targeted misclassification or reduced benchmark
accuracy~\citep{zhao2023evaluating, schlarmann2023adversarial},
hallucinated descriptions~\citep{li2023pope}, manipulated answer
preferences~\citep{lan2025phi}, or safety
bypass~\citep{qi2024visual}.
Only a few works study vulnerability on model allocation mechanisms. For
example, attacks on LLM routers show that adversarial inputs can
manipulate which model is selected while preserving response
quality~\citep{shafran2025rerouting, lin2025life}. These studies show
that multi-model systems introduce a control-flow attack surface in
addition to the prediction surface of individual models. Routing
attacks typically select the model before generation, while cascades
make the allocation decision after the weak model has generated an
answer and produced a confidence signal.

\begin{figure}[ht]
  \centering
  \includegraphics[width=\linewidth]{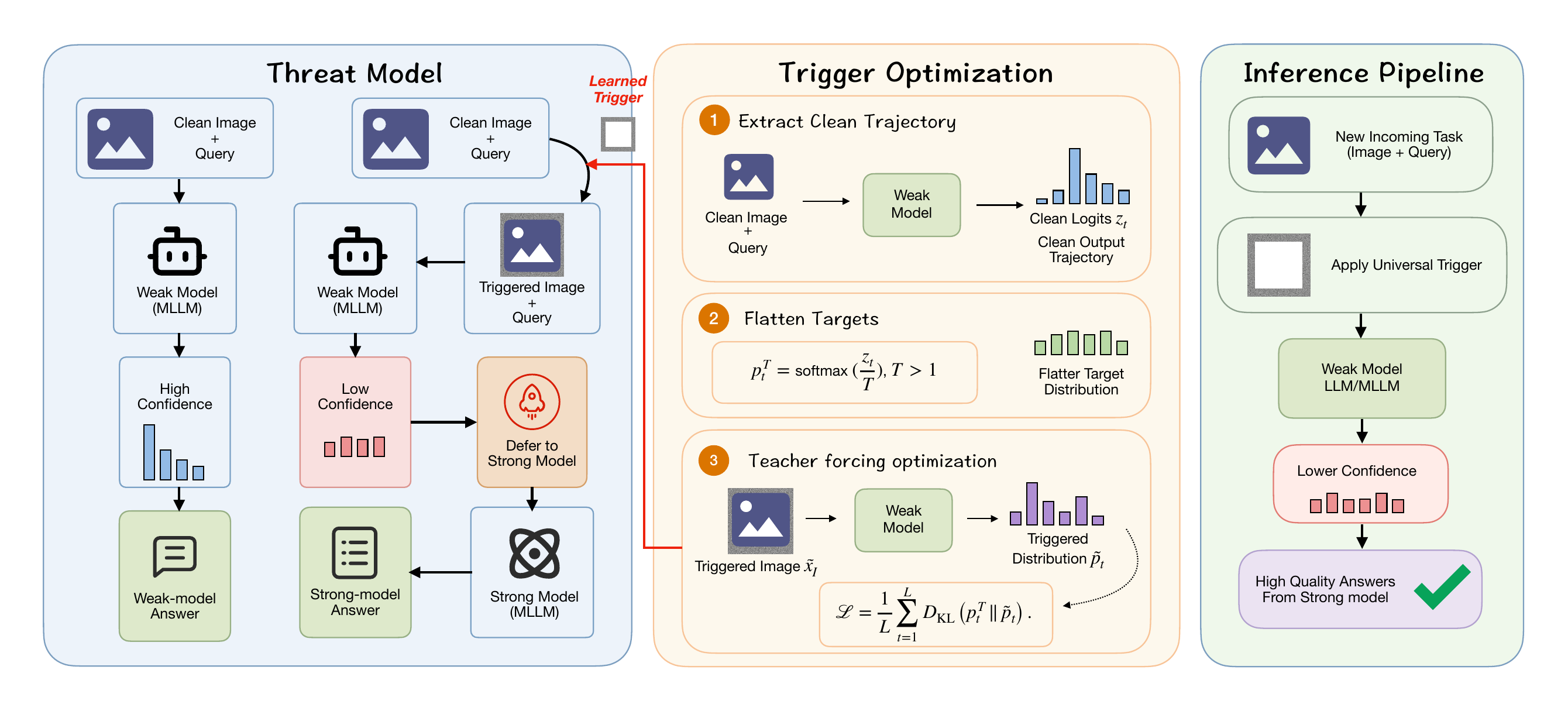}
  \caption{Overview of the forced deferral attack on MLLM cascades. The left panel illustrates the standard cascade pipeline, where a weak model answers high-confidence inputs directly and only defers uncertain cases to a stronger model. The middle panel shows our attack procedure: we first extract the clean output trajectory, construct flattened temperature-scaled target distributions, and optimize a learned border trigger using teacher-forced KL minimization. The right panel shows the deployment effect: applying the learned trigger lowers the weak model’s confidence, causing more inputs to be routed to the strong model while preserving high-quality final answers.}
  \label{fig:overview}
\end{figure}

\section{Method}
\subsection{Preliminary on Uncertainty-Based MLLM Cascade}
Following ~\citep{chen2023frugalgpt, gupta2024languagemodelcascadestokenlevel}, we consider a service provider that deploys an MLLM cascade to reduce inference cost, as illustrated in Figure~\ref{fig:overview}. The cascade consists of a cheap but weaker model $M_w$, and an expensive but stronger model $M_s$. Given an image--question pair $(x_I, x_Q)$, the system first queries the weak model and obtains its response $y_w = M_w(x_I, x_Q)$. It then computes a post-hoc confidence score $c_w$ from the weak model's generation. The confidence score may be derived from answer-token probabilities, sequence likelihood, perplexity, output consistency, or other uncertainty quantification metrics.

The cascade uses this confidence score to decide whether to accept the weak model's answer or defer the query to the strong model. Formally, given a fixed threshold $\gamma$, the final response $y_{\mathrm{cas}}$ is
\begin{equation}
y_{\mathrm{cas}}(x_I,x_Q)=
\begin{cases}
y_w, & c_w \geq \gamma,\\
M_s(x_I,x_Q), & c_w < \gamma.
\end{cases}
\end{equation}
The threshold $\gamma$ is selected before deployment, for example by choosing a target deferral rate on clean validation data. Under benign inputs, the provider expects easy queries to be answered by the weak model $M_w$, while only difficult or uncertain queries are routed to the stronger model $M_s$. Therefore, the weak model's confidence acts as a computation-allocation signal: lowering $c_w$ increases the likelihood of invoking the stronger and more expensive model. This exposes a new vulnerability: by manipulating $c_w$, an attacker can force the cascade to defer beyond its intended operating point, which we formalize in the next section.

\subsection{Threat Model in Deferral Attack}
\label{sec:threat-model}

We propose an adversarial image attack that manipulates the deferral decision of an MLLM cascade. 

\textbf{Attacker Goal:}
Given a clean input $(x_I, x_Q)$, the attacker constructs $(\tilde{x}_I(\delta), x_Q)$, where $\delta$ denotes an image space perturbation and $\tilde{x}_I(\delta)$ denotes the perturbed image. The attack modifies only the image while leaving $x_Q$ unchanged. The attack succeeds when $c_w(\tilde{x}_I, x_Q) < \gamma$, so the cascade routes the query to $M_s$. The attacker further requires $\tilde{x}_I$ to remain answerable by $M_s$, since the attacker aims at a high-quality answer.
\textbf{Attacker Knowledge:}
The attacker can access the weak model's logits during attack construction (e.g., when $M_w$ is open-source, or through a surrogate from the same family), but does not know the cascade's uncertainty metric $c_w(\cdot)$ or threshold $\gamma$. The attack therefore cannot rely on optimizing any specific confidence score.
\textbf{Attacker Capability:}
The perturbation $\delta$ is learned once on a training set and applied to unseen inputs at deployment time, so the per-query cost of the attack is negligible.

\subsection{Practical Principles for Designing Deferral Attack}

We design the image-space perturbation $\delta$ around three principles derived from the threat model: the perturbation should be \emph{metric-agnostic}, \emph{semantics-preserving}, and \emph{universal across queries}. We describe each principle below, then introduce a training objective and trigger parameterization that satisfy all three:
\begin{itemize}[leftmargin=*]
    \item \textbf{Metric-agnostic.} Because the router's uncertainty signal is unknown, optimizing against any particular score (such as maximum softmax probability, predictive entropy, or the margin between the top two tokens) is unlikely to transfer to a different score, and would be invalidated by any change of metric or threshold on the operator's side. We must therefore target the underlying distributional property that all such metrics reflect, rather than any single one of them.
    \item \textbf{Semantics-preserving.} The triggered query is precisely what the strong model receives once deferral succeeds. If the perturbation degrades the visual content needed to answer the question, the strong model also fails and the adversary gains nothing from the deferral. The perturbation must therefore preserve necessary semantic content.
    \item \textbf{Universal across Queries.} Reoptimizing for each new query is too expensive, since the cost would scale with the number of attack queries. The trigger should instead be optimized once and reused across inputs, making the optimization a one-time cost shared across the attack.

\end{itemize}

\subsection{Attack Implementation}
We address these requirements through two design components. We first introduce a training objective that handles metric-agnosticism by acting on the weak model's token distributions rather than any specific score. We then apply this objective to optimize a structural transformation for an adversarially perturbed image input $\tilde{x}_I(\delta)$, which preserves the central visual content for semantics-preservation and is optimized once and applied universally across queries.

The key question is how to increase uncertainty in a way that is independent of the router's specific confidence metric. A natural starting point is the weak model's next-token distribution: when this distribution is sharply concentrated, many confidence scores, including maximum token probability, entropy, perplexity, and margin-based scores, indicate high confidence. Thus, a metric-agnostic attack should make this distribution less concentrated. For a training sample $(x_I,x_Q)$, let $y=(y_1,\ldots,y_L)$ be the weak model's greedy response on the clean input. During training, we use this clean response as the reference sequence: at step $t$, the weak model is conditioned on the perturbed image $\tilde{x}_I(\delta)$, the query $x_Q$, and the clean prefix $y_{<t}$, producing the next-token distribution $p_w(\cdot \mid (\tilde{x}_I(\delta), x_Q), y_{<t})$.

The most direct way to increase uncertainty would be to maximize the entropy of this distribution. Equivalently, this minimizes its KL divergence to the uniform distribution $u(\cdot)=1/|\mathcal{V}|$:
\begin{equation}
\max_{\delta}\; H\!\left(p_w(\cdot \mid (\tilde{x}_I(\delta), x_Q), y_{<t})\right)
\quad \Longleftrightarrow \quad
\min_{\delta}\; D_{\mathrm{KL}}\!\left(
p_w(\cdot \mid (\tilde{x}_I(\delta), x_Q), y_{<t}) \,\|\, u(\cdot)
\right),
\end{equation}
since $D_{\mathrm{KL}}\!\left(p_w(\cdot \mid (\tilde{x}_I(\delta), x_Q), y_{<t}) \,\|\, u(\cdot)\right)=\log |\mathcal{V}|-H\!\left(p_w(\cdot \mid (\tilde{x}_I(\delta), x_Q), y_{<t})\right)$.

However, directly driving the weak model toward a uniform distribution discards which tokens the clean model considered plausible. This can move the prediction away from the clean response and degrade the quality of the answer produced by the strong model. Therefore, the perturbation should reduce confidence while preserving the information needed to answer the original query.

We therefore replace the uniform target with a flattened version of the weak model's clean distribution. At step $t$, using the same clean prefix $y_{<t}$, let $z_t = M_w(x_I, x_Q, y_{<t})$ denote the clean logits. We define the flattened target
\begin{equation}
q_t^T(\cdot) = \mathrm{softmax}(z_t/T), \qquad T>1.
\end{equation}
Temperature scaling reduces concentration while preserving the ranking of tokens under the clean logits. The target therefore increases uncertainty while keeping the optimization anchored to the weak model's original response, making it less likely to degrade the answer quality after deferral to the strong model. With this revised target, we optimize the perturbation by matching the weak model's next-token distribution on the perturbed image to the flattened clean distribution:
\begin{equation}
\delta^\star =
\arg\min_{\delta}
\frac{1}{|\mathcal{D}|}
\sum_{(x_I,x_Q)\in\mathcal{D}}
\frac{1}{L}
\sum_{t=1}^{L}
D_{\mathrm{KL}}\!\left(
q_t^T(\cdot)
\,\|\,
p_w(\cdot \mid (\tilde{x}_I(\delta), x_Q), y_{<t})
\right).
\end{equation}
This objective lowers confidence by flattening the weak model's distribution rather than forcing a specific incorrect answer. Because the target preserves the clean token ranking, it provides a more stable signal than a uniform target while remaining agnostic to the cascade's actual uncertainty metric.

\paragraph{Universal Border Trigger}

The content needed to answer a VQA question can appear anywhere in the image. A naive global perturbation would touch this content directly, while a single corner patch would only mask one side and could be entirely outside the relevant region for some samples but inside it for others. A frame-shaped border surrounds the image symmetrically and stays clear of the central content regardless of where the question's referent appears.

To preserve the input resolution expected by the MLLM, the clean image is resized into the inner region and the border is filled with the trigger, so that $\delta$ has a dedicated set of pixels that does not overlap with the original content. Let $b$ be a binary mask that is $1$ on the border and $0$ on the inner region, and let $r(x_I)$ denote the clean image resized into the inner region. The triggered image is
\begin{equation}
\tilde{x}_I(\delta) = b \odot \delta + (1 - b) \odot r(x_I),
\end{equation}
where $\odot$ denotes elementwise multiplication. The border width trades off attack strength against the utility on the strong model. We use a fixed width across experiments and study its effect in \S\ref{sec:ablation}.

\section{Experiments}

This section evaluates the effectiveness and generalizability of the proposed deferral attack in manipulating MLLM cascades. Specifically, we aim to answer the following research questions:
\begin{itemize}[leftmargin=*]
    \item \textbf{Q1:} Can FDA effectively force the cascade to reroute perturbed inputs to the strong model?
    \item \textbf{Q2:} Will the attacker receive compelling end-to-end system performance with the perturbed inputs?
    \item \textbf{Q3:} Does FDA generalize across uncertainty metrics, datasets, and model families?
\end{itemize}

We give positive answers to Q1 and Q2, by comparing cascading reroute rate, system accuracy and weak model confidence of FDA against generic image perturbation and prompt injection baselines. Q3 is answered through study on cross-dataset/family, preprocessing defenses, and ablation variants.

\subsection{Experimental Setup}

\textbf{Datasets and Models.}
We evaluate on three multimodal benchmarks, MMBench~\citep{liu2024mmbench}, ScienceQA~\citep{lu2022learn}, and MMMU~\citep{yue2024mmmu}. We focus on image-based multiple choice questions, where the model outputs an answer option followed by a brief explanation. Details on dataset construction are provided in Appendix~\ref{app:experimental_details}.

We evaluate the MLLM cascade system based on two representative families: Gemma4-E2B/26B~\citep{gemma4modelcard} and Qwen3-VL-2B/32B~\citep{bai2025qwen3}, each with a weak-strong model pair. The evaluation considers cascading among both same-family and cross-family models.

\textbf{Cascade Setting.} We instantiate the uncertainty-based cascade described above with three weak-model confidence signals: \textit{Answer Probability (AP)}~\citep{mahaut2024factual}, the probability assigned to the generated answer token; \textit{Inverse Perplexity ($1/\mathrm{PPL}$)}~\citep{fomicheva2020unsupervised, lin2023generating}, the inverse of the standard perplexity over the generated response, equivalently the geometric mean of token probabilities; and \textit{Jaccard Similarity (JS)}~\citep{lin2023generating}, the mean pairwise word-level Jaccard similarity across multiple sampled responses. Across three measurements, a larger value indicates a higher weak-model confidence and certainty. Following standard practice in selective inference and
cascading~\citep{chen2023frugalgpt, aggarwal2023automix},
we calibrate a separate threshold for each confidence score on the
clean training split so that $\rho$ fraction of the lowest-confidence queries are
deferred. This threshold is then fixed and applied to both clean and
attack-triggered test inputs.

\begin{wraptable}{r}{0.56\textwidth}
\vspace{-1.2em}
\caption{Mean confidence scores on clean and attacked test samples. Each cell reports Clean/Attacked (with confidence change $\Delta=\mathrm{Attacked}-\mathrm{Clean}$).}
\vspace{-0.3em}
\centering
\scriptsize
\setlength{\tabcolsep}{2.0pt}
\renewcommand{\arraystretch}{1.05}
\resizebox{\linewidth}{!}{
\begin{tabular}{@{}ccccc@{}}
\toprule
\multirow{2}{*}{Model} & \multirow{2}{*}{Metric $\downarrow$} & \multicolumn{3}{c}{Clean/Attacked ($\Delta$)} \\
\cmidrule(lr){3-5}
& & MMBench & MMMU & ScienceQA \\
\midrule
\multirow{3}{*}{Gemma}
& AP    & 0.94/0.24 $(-.70)$ & 0.87/0.04 $(-.83)$ & 0.96/0.10 $(-.86)$ \\
& 1/PPL & 0.87/0.51 $(-.36)$ & 0.86/0.49 $(-.37)$ & 0.90/0.42 $(-.48)$ \\
& JS    & 0.61/0.33 $(-.28)$ & 0.55/0.30 $(-.26)$ & 0.67/0.32 $(-.35)$ \\
\midrule
\multirow{3}{*}{Qwen}
& AP    & 0.94/0.03 $(-.91)$ & 0.73/0.09 $(-.65)$ & 0.94/0.27 $(-.66)$ \\
& 1/PPL & 0.83/0.52 $(-.30)$ & 0.80/0.59 $(-.21)$ & 0.84/0.61 $(-.23)$ \\
& JS    & 0.58/0.22 $(-.36)$ & 0.48/0.24 $(-.24)$ & 0.56/0.31 $(-.25)$ \\
\bottomrule
\end{tabular}
}
\label{tab:uq_shift}
\end{wraptable}

\textbf{Attack Settings, Baselines and Metrics.} The optimized universal trigger is placed on the image border with width 16 pixels. We optimize the flattened objective with temperature $T\in\{3,4\}$ and select the
final temperature based on validation performance; details are provided
in Appendix~\ref{app:experimental_details}. Since this is the first attack targeting MLLM cascading system, we compare our attack with two crafted baselines:  \textit{defocus-blur perturbation}~\citep{hendrycks2019benchmarking} representing generic image perturbations, and \textit{prompt-injection} modifying the textual query to question the model itself. The full prompt used for prompt injection and the exact perturbation parameters are provided in Appendix~\ref{app:experimental_details}. We report two main attack evaluation metrics: 1) \textbf{Reroute Rate}, defined as the fraction of test samples routed to the strong model under the fixed calibrated deferral threshold. This directly measures the effectiveness of the attack in manipulating cascading decisions; 2) \textbf{System Accuracy}, defined as the end-to-end accuracy of the cascade after routing decisions are made: samples not deferred are answered by the weak model, while deferred samples are answered by the strong model. This metric captures whether the increased routing still leads to desired system-level performance for the benefit of the attacker, rather than simply corrupting the input.

\begin{figure}[t]
  \centering

  \begin{subfigure}{0.9\linewidth}
    \centering
    \includegraphics[
      width=\linewidth,
      trim=0 10 0 0,
      clip
    ]{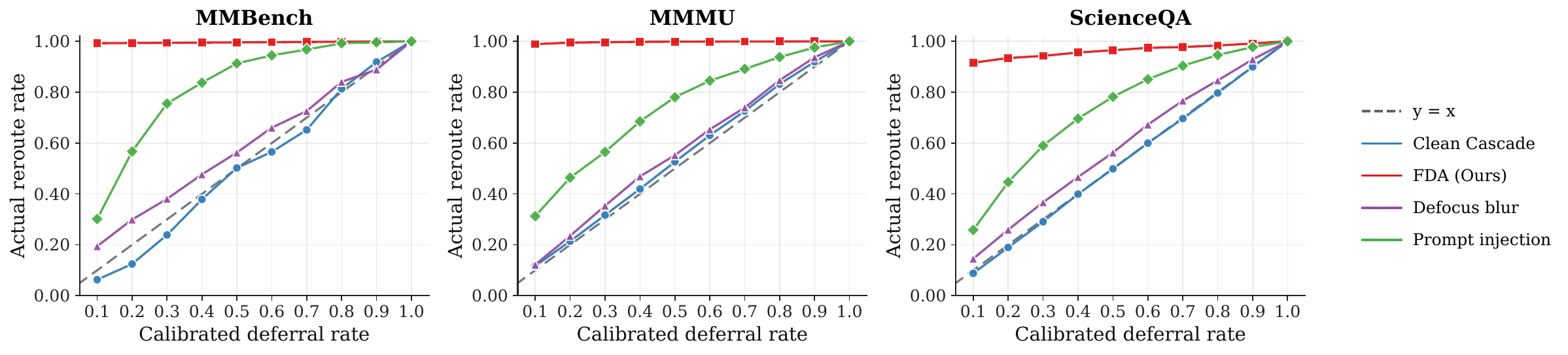}
    \caption{Qwen3-VL-2B$\rightarrow$Qwen3-VL-32B cascade system under JS metric.}
    \label{fig:reroute_qwen}
  \end{subfigure}

  \begin{subfigure}{0.9\linewidth}
    \centering
    \includegraphics[
      width=\linewidth,
      trim=0 10 0 0,
      clip
    ]{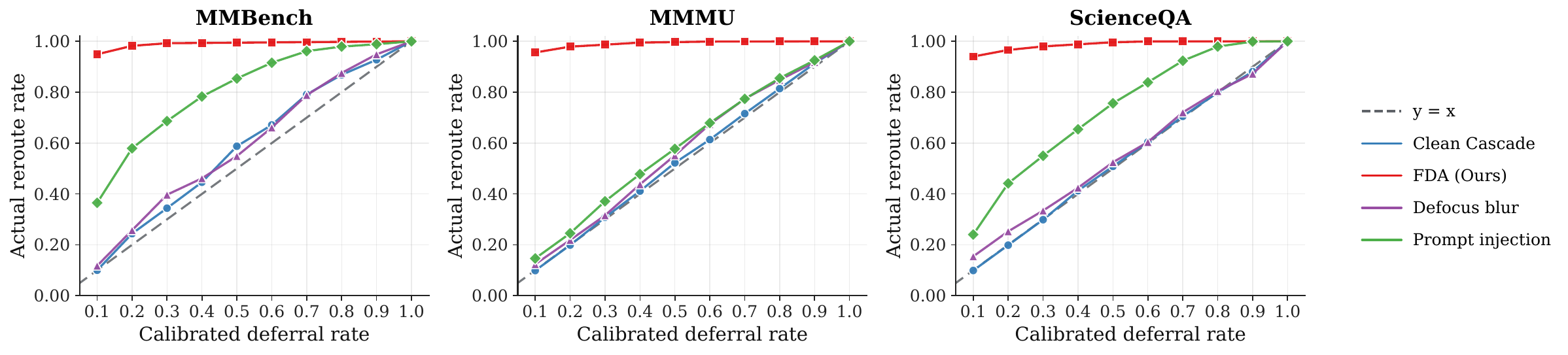}
    \caption{Gemma4-E2B$\rightarrow$Gemma4-26B cascade system under 1/PPL metric.}
    \label{fig:reroute_gemma4}
  \end{subfigure}

  \caption{Reroute rate under varying calibrated deferral rate, with dashed line indicating accurate calibration.}
  \label{fig:reroute_main}
\end{figure}

\begin{figure}[ht]
  \centering
  \includegraphics[
    width=0.9\linewidth,
    trim=0 10 0 0,
    clip
  ]{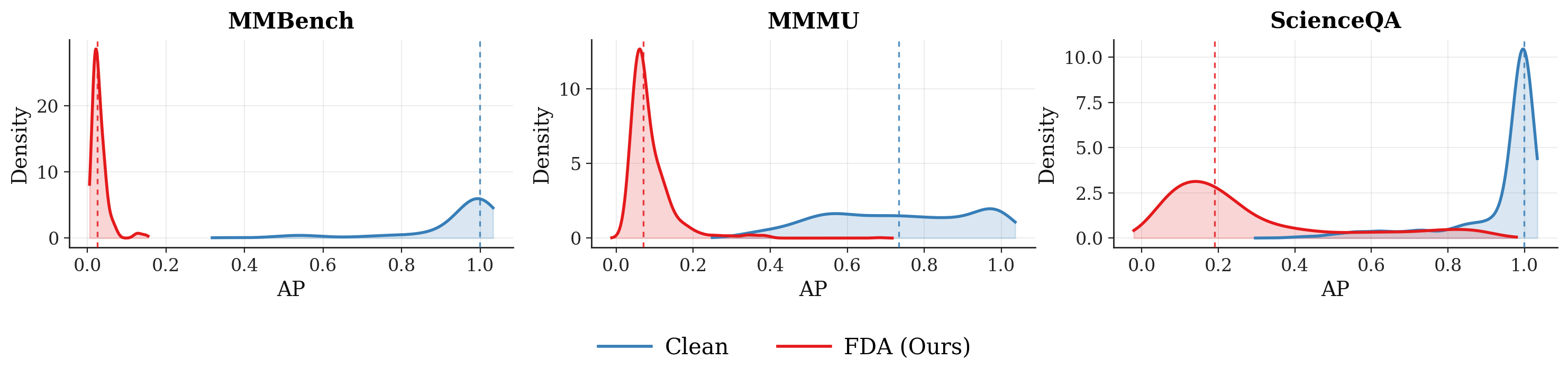}
 \caption{AP distributions on clean and attacked test inputs for the Qwen3-VL weak model across three datasets, estimated by Gaussian KDE. Dashed vertical lines mark the distribution medians.}
  \label{fig:answer_prob_shift}
\end{figure}

\subsection{Main Results on Attack Reroute Rate}
\label{sec:main_results}

This subsection evaluates whether FDA can directly manipulate the routing decision of an MLLM cascade. Since the goal of the attack is to make more inputs routed to the strong model, we use \textbf{reroute rate} as the primary metric. Higher reroute rates indicate stronger attack effectiveness. 

Figures~\ref{fig:reroute_qwen} and~\ref{fig:reroute_gemma4} show representative rerouting results for two same family cascades: Qwen under JS and Gemma under $1/\mathrm{PPL}$. Additional results under other uncertainty metrics are provided in Appendix~\ref{app:additional_results}.
In the figures, a successful deferral attack should push the curve well above the clean cascade and close to one, even when the calibrated clean deferral rate is small.

\textbf{FDA consistently drives routing to the strong model.} Across MMBench, MMMU, and ScienceQA, FDA pushes the actual reroute rate close to one over most calibrated deferral rates, especially when the clean deferral rate is small. In contrast, defocus blur stays near the clean cascade, and prompt injection increases rerouting only moderately. These results show that FDA reliably manipulates cascade routing under fixed calibrated thresholds.

\subsection{Trigger-Induced Confidence Shift on Weak Model}

\begin{wrapfigure}{r}{0.48\linewidth}
  \vspace{-1.0em}
  \centering
  \includegraphics[width=\linewidth]{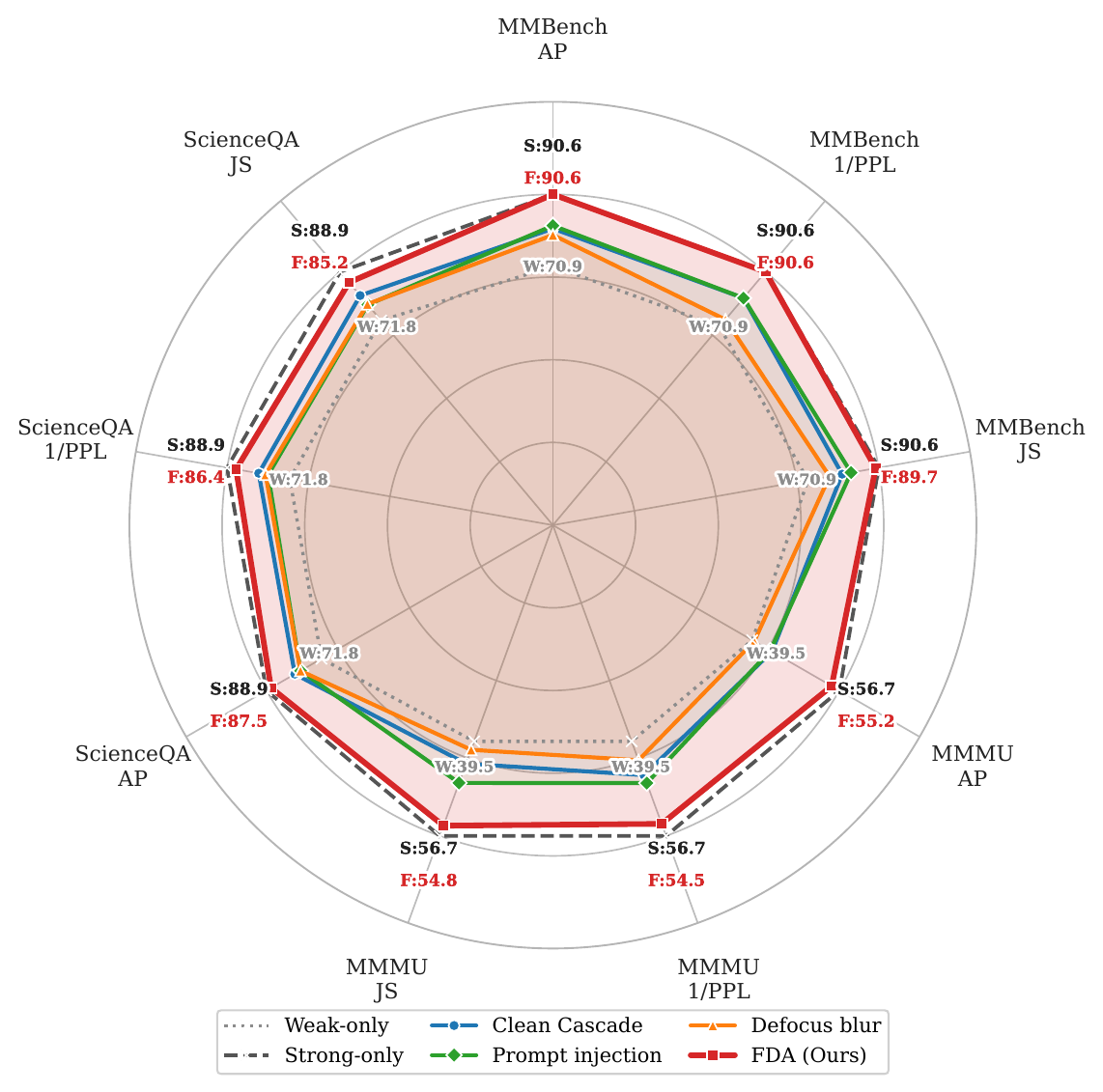}
  \caption{Cascade performance under the FDA attack for the Gemma same-family cascade at calibrated deferral rate $D=0.3$.}
  \label{fig:System_acc_gemma}
\end{wrapfigure}

This subsection examines why FDA increases rerouting by analyzing how it changes the weak model's confidence scores. We compare AP, 1/PPL, and JS on clean and attacked test samples, where lower values indicate lower weak-model confidence. Table~\ref{tab:uq_shift} summarizes this effect by reporting the mean value of each confidence score before and after attack. Figure~\ref{fig:answer_prob_shift} complements the table by showing the full per-sample AP distributions: for each dataset, we estimate the density of clean and attacked AP values using Gaussian KDE, so the figure shows whether the attack shifts the entire sample distribution.

Table~\ref{tab:uq_shift} shows that FDA consistently lowers confidence across both weak models, all datasets, and all confidence scores. The drop is largest for AP, but 1/PPL and JS also decrease, suggesting that the trigger affects multiple confidence signals rather than only one metric. Figure~\ref{fig:answer_prob_shift} shows the same trend at the distribution level: attacked samples shift toward lower AP values, so more of them fall below the fixed threshold calibrated on clean inputs and are routed to the strong model.

\subsection{Cascade Performance Gain from Deferral Attack}

We next examine whether forced deferral translates into end-to-end cascade performance gains. Figure~\ref{fig:System_acc_gemma} reports system accuracy for the Gemma same-family cascade at calibrated deferral rate $D=0.3$ across three datasets and three uncertainty metrics. We compare the clean cascade, FDA, two baselines, and the weak-only and strong-only reference accuracies.

FDA consistently moves cascade performance toward the strong-only reference. On MMBench, it nearly matches strong-only accuracy across metrics, reaching about $90\%$ system accuracy. On MMMU and ScienceQA, FDA also improves over the clean cascade and both baselines, closing much of the gap to strong-only performance. These results show that forced deferral yields real system-level gains, not merely a change in routing.

\subsection{Generalization Across Datasets and Model Families}

We further test whether FDA remains useful when moving beyond the original evaluation setting. We consider two cases: routing attacked inputs to a strong model from another family, and applying a trigger trained on one dataset to different datasets.

\begin{wraptable}{r}{0.55\linewidth}
\vspace{-1.0em}
\caption{Cross-family cascade performance under AP. Clean columns report the clean cascade accuracy. FDA columns report Same FDA / Cross FDA accuracy, with $\Delta=\text{Cross}-\text{Same}$ in parentheses.}
\vspace{-0.4em}
\centering
\scriptsize
\setlength{\tabcolsep}{3pt}
\resizebox{\linewidth}{!}{
\begin{tabular}{cccccc}
\toprule
\multirow{2}{*}{Weak model} & \multirow{2}{*}{Dataset}
& \multicolumn{2}{c}{Clean}
& \multicolumn{2}{c}{Same FDA / Cross FDA $(\Delta)$} \\
\cmidrule(lr){3-4} \cmidrule(l){5-6}
& & $D=0.1$ & $D=0.3$ & $D=0.1$ & $D=0.3$ \\
\midrule
\multirow{3}{*}{Gemma4-E2B}
& MMBench   & 0.744 & 0.812 & 0.897 / 0.880 {\scriptsize $(-.017)$} & 0.906 / 0.889 {\scriptsize $(-.017)$} \\
& MMMU      & 0.402 & 0.434 & 0.551 / 0.591 {\scriptsize $(+.041)$} & 0.552 / 0.592 {\scriptsize $(+.041)$} \\
& ScienceQA & 0.748 & 0.801 & 0.873 / 0.956 {\scriptsize $(+.083)$} & 0.875 / 0.959 {\scriptsize $(+.084)$} \\
\midrule
\multirow{3}{*}{Qwen3-VL-2B}
& MMBench   & 0.802 & 0.855 & 0.889 / 0.906 {\scriptsize $(+.017)$} & 0.889 / 0.906 {\scriptsize $(+.017)$} \\
& MMMU      & 0.420 & 0.479 & 0.597 / 0.567 {\scriptsize $(-.030)$} & 0.597 / 0.567 {\scriptsize $(-.030)$} \\
& ScienceQA & 0.884 & 0.938 & 0.949 / 0.876 {\scriptsize $(-.073)$} & 0.951 / 0.878 {\scriptsize $(-.073)$} \\
\bottomrule
\end{tabular}
}
\label{tab:cross_family_performance}
\vspace{-1.0em}
\end{wraptable}

\textbf{Cross-family Cascades.}
Since the weak model is unchanged, FDA still successfully induces deferral. Table~\ref{tab:cross_family_performance} therefore focuses on whether the rerouted strong model can preserve overall accuracy. Cross-family FDA improves over the clean cascade in all cases, and its performance is generally close to same-family FDA, with small gains or drops depending on the selected strong model.

\begin{figure}[ht]
  \centering
  \includegraphics[width=0.8\linewidth]{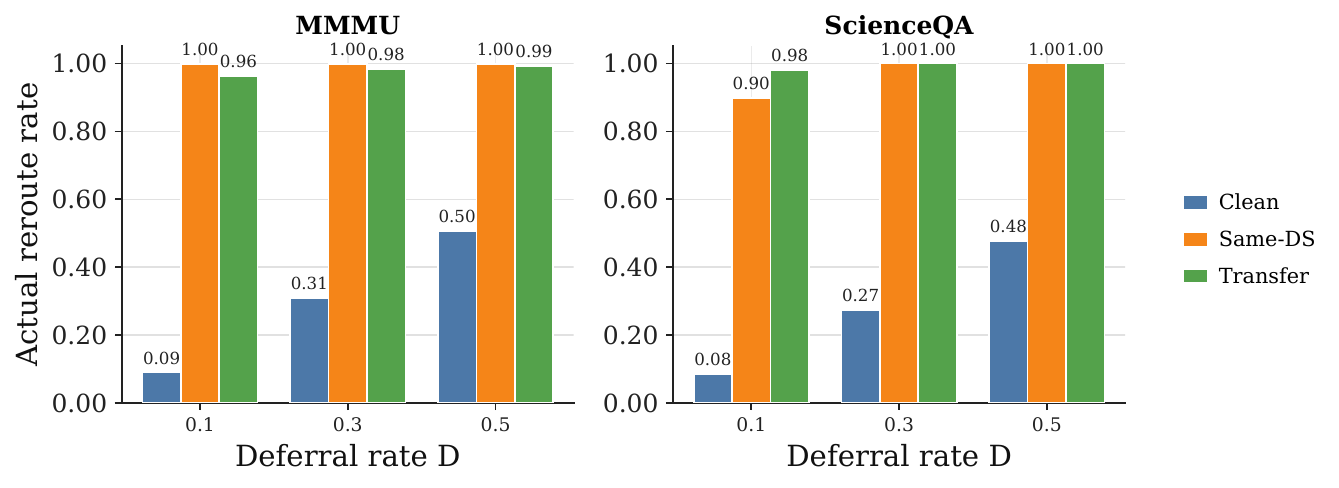}
 \caption{Cross-dataset transferability of FDA from MMBench to MMMU and ScienceQA using Qwen3-VL-2B under AP. Bars show actual reroute rates at $D\in\{0.1,0.3,0.5\}$.}
  \label{fig:transfer_atp_qwen}
\end{figure}

\textbf{Cross-dataset Transferability.}
Figure~\ref{fig:transfer_atp_qwen} shows that an FDA trigger optimized on MMBench transfers well to MMMU and ScienceQA. The transferred trigger achieves near-complete rerouting on both target datasets, closely matching the same-dataset FDA trigger across calibrated deferral rates. This suggests that FDA learns a transferable confidence-suppression pattern rather than memorizing dataset-specific features.

\subsection{Attack Resilience to Preprocessing Defenses}

We test whether simple preprocessing can remove the FDA trigger before routing. We consider JPEG compression, Gaussian noise, and 5-bit color reduction. FDA remains effective under all three defenses: bit-depth reduction has little effect, Gaussian noise only weakens rerouting at low deferral rates, and JPEG is the strongest but still does not restore clean-cascade behavior. Thus, common preprocessing defenses are insufficient to reliably remove FDA's rerouting effect.

\begin{figure}[ht]
  \centering
  \includegraphics[
    width=0.9\linewidth,
    trim=0 10 0 0,
    clip
  ]{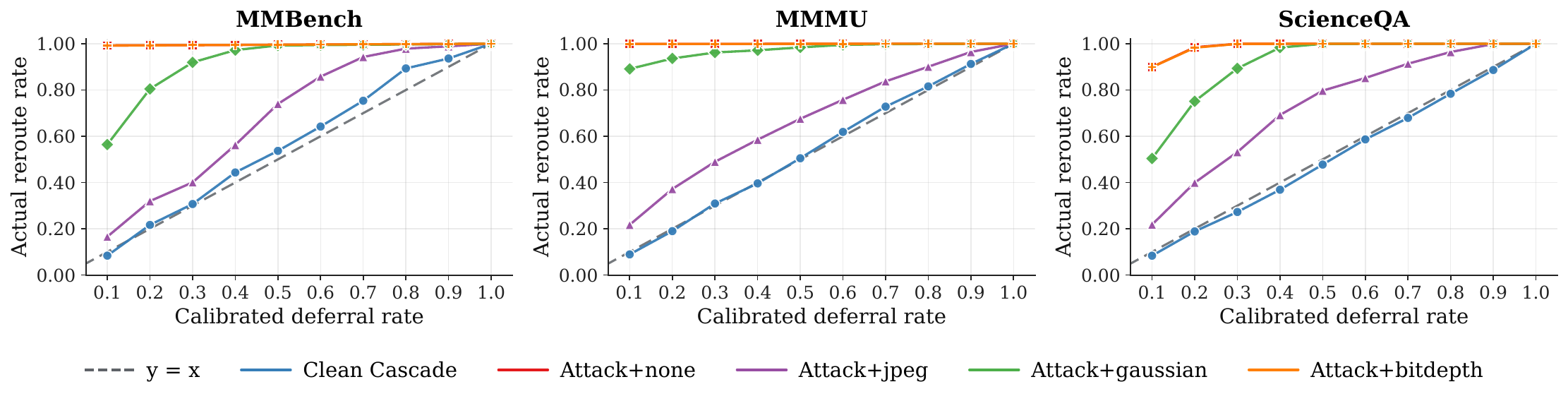}
 \caption{Effect of preprocessing defenses on FDA under AP. Results are shown for the Qwen-to-Qwen cascade across MMBench, MMMU, and ScienceQA. We compare FDA without preprocessing to FDA after JPEG compression, Gaussian noise, and bit depth reduction.}
  \label{fig:defense_atp_reroute}
\end{figure}

\subsection{Ablation Study}
\label{sec:ablation}

We conduct ablation studies on the Gemma family cascade using MMBench. Table~\ref{tab:ablation} reports system accuracy at two representative calibrated deferral rates, $D=0.1$ and $D=0.5$, under AP, PPL, and JS. Overall, the default setting, using a 16-pixel border, $T=3$, and the Border+Temp-KL objective, gives the best or near-best accuracy across metrics and deferral rates.

\begin{table*}[ht]
\caption{Ablation study on Gemma4-E2B and MMBench. We report system accuracy at calibrated deferral rates $D=0.1$ and $D=0.5$. Default settings are marked with $^\star$. Temp-KL denotes the temperature flattened KL objective.}
\vspace{0.3em}
\centering
\scriptsize
\setlength{\tabcolsep}{2.5pt}
\resizebox{\textwidth}{!}{
\begin{tabular}{llc ccccc ccccc ccc}
\toprule
& & \multicolumn{1}{c}{Baseline} & \multicolumn{5}{c}{Border width} & \multicolumn{5}{c}{Temperature} & \multicolumn{3}{c}{Objective / Perturbation Variant} \\
\cmidrule(lr){3-3} \cmidrule(lr){4-8} \cmidrule(lr){9-13} \cmidrule(l){14-16}
Metric & $D$ & Clean & bw=4 & bw=8 & bw=16$^\star$ & bw=32 & bw=48 & T=1 & T=2 & T=3$^\star$ & T=4 & T=5 & Border+Temp-KL$^\star$ & Global+Temp-KL & Border+MaxEnt \\
\midrule
\multirow{2}{*}{AP}
& 0.1 & 0.7436 & 0.8889 & 0.8693 & \textbf{0.8974} & 0.8718 & 0.8632 & 0.7472 & 0.7488 & \textbf{0.8974} & 0.8803 & 0.8845 & \textbf{0.8974} & 0.8632 & 0.7265 \\
& 0.5 & 0.8718 & \textbf{0.9231} & 0.8889 & 0.9060 & 0.8718 & 0.8632 & 0.8718 & 0.8718 & 0.9060 & 0.8889 & 0.8889 & 0.9060 & 0.8974 & 0.8419 \\
\midrule
\multirow{2}{*}{PPL}
& 0.1 & 0.7436 & 0.8547 & 0.7874 & \textbf{0.8974} & 0.8718 & 0.8597 & 0.7350 & 0.7436 & \textbf{0.8974} & 0.8547 & 0.8395 & \textbf{0.8974} & 0.8462 & 0.7350 \\
& 0.5 & 0.8632 & \textbf{0.9060} & 0.8803 & \textbf{0.9060} & 0.8718 & 0.8632 & 0.8632 & 0.8291 & \textbf{0.9060} & 0.8889 & 0.8889 & \textbf{0.9060} & 0.8889 & 0.8314 \\
\midrule
\multirow{2}{*}{JS}
& 0.1 & 0.7436 & 0.8718 & 0.8718 & \textbf{0.8868} & 0.8291 & 0.8291 & 0.7521 & 0.7199 & \textbf{0.8868} & 0.8803 & 0.8755 & \textbf{0.8868} & 0.7521 & 0.7436 \\
& 0.5 & 0.8659 & 0.8889 & 0.8889 & \textbf{0.9060} & 0.8521 & 0.8632 & 0.8632 & 0.8547 & \textbf{0.9060} & 0.8824 & 0.8889 & \textbf{0.9060} & 0.8632 & 0.8376 \\
\bottomrule
\end{tabular}
}
\label{tab:ablation}
\end{table*}

\textbf{Border Width.}
The 16-pixel border gives the most stable performance. Smaller borders can be less effective, while larger borders tend to reduce accuracy, likely because they occupy more visual space. This suggests that the border should be large enough to affect the weak model, but not so large that it harms the visual content used by the strong model.

\textbf{Temperature.}
Low temperatures, such as $T=1$ and $T=2$, lead to weaker performance because the target distributions remain close to the clean distributions. Increasing the temperature improves the attack effect, but higher temperatures do not further improve accuracy. The default $T=3$ gives the strongest overall results across AP, PPL, and JS.

\textbf{Objective and Perturbation.}
Border+Temp-KL performs better than both Global+Temp-KL and Border+MaxEnt in most settings. This shows that both parts are important: Temp-KL provides a more effective confidence flattening objective, while placing the perturbation on the border helps preserve the image content needed by the strong model.

\section{Conclusion and Limitations}

In this paper, we study confidence manipulation in MLLM cascades. We introduce the Forced Deferral Attack (FDA), which lowers the weak model's confidence and reroutes more queries to the strong model. FDA learns a universal border trigger with a temperature-flattened objective that reduces distributional concentration while preserving answerability. Across datasets, metrics, and model families, FDA substantially increases strong-model routing, transfers across settings, and remains effective under common preprocessing defenses. These results show that cascades can be attacked at the compute-allocation level, weakening their intended efficiency gains.

This study also points to broader open challenges for robust cascaded inference. Confidence-based routing may behave differently across deployed systems, open-ended tasks, and multi-turn interactions, and the space of possible allocation attacks extends beyond the border-trigger design studied here. Designing confidence estimates and routing policies that remain reliable under adversarial input transformations remains an important unresolved problem.

\clearpage
{\small
\bibliographystyle{unsrtnat}
\bibliography{references}
}

\clearpage
\appendix

\section{Experimental Details}
\label{app:experimental_details}

\subsection{Dataset Filtering and Split Construction}
\label{app:dataset_details}

We evaluate on MMBench, ScienceQA, and MMMU. For all datasets, we keep only image based multiple choice questions, so that each sample contains a visual input and answer correctness can be evaluated automatically from the predicted option.

For MMBench, we use the development set. After filtering, we randomly split the selected samples into training, validation, and test subsets with an 8:1:1 ratio. For MMMU, we use the test set and apply the same 8:1:1 split after filtering. For ScienceQA, we use the official training, validation, and test splits, and filter each split to keep only samples that contain both an image and a multiple choice question.

For MMMU, we additionally require each selected sample to contain exactly one image, non-empty question text, and text-only answer options. Samples whose options contain image references or cannot be mapped to a discrete multiple choice answer are removed. For ScienceQA, when a hint is provided, we include it as part of the textual question context. All random splits use a fixed random seed.

\subsection{Prompt Format and Answer Extraction}

For all datasets, each input contains an image and a text prompt. We use a unified multiple choice prompt format, where the valid option letters are generated according to the number of options in the current sample:
\begin{quote}
\{question\}

A. \{option\_A\}

B. \{option\_B\}

C. \{option\_C\}

\...

\{letter\_K\}. \{option\_K\}

Answer with \{valid\_option\_letters\} only on the first line.

On the second line, briefly explain your choice.
\end{quote}
Here, $K$ is the number of answer options for the current sample, $\{letter\_K\}$ is the last valid option letter, and $\{valid\_option\_letters\}$ is rendered as a natural language list such as ``A, B, C, or D'' or ``A, B, C, D, or E''. We wrap the text prompt and image content using the model processor's chat template and set \texttt{add\_generation\_prompt=True}. Unless otherwise specified, we use greedy decoding with \texttt{do\_sample=False} and \texttt{max\_new\_tokens=64}. We use the answer option on the first line for accuracy evaluation and answer token probability, and use the full generated response to compute sequence based metrics such as perplexity and Jaccard.

To extract the final answer, we decode the generated token sequence into a string and take the first non-empty line. We first match an isolated answer option using a regular expression over the valid option letters for the current question. If this fails, we match any valid option letter in the first line. If the extracted letter belongs to the valid option set for the question, we use it as the predicted answer; otherwise, the prediction is treated as invalid.

\subsection{Dataset Filtering and Splits}

Table~\ref{tab:dataset_filtering} summarizes the source split, filtering rule, and final train, validation, and test sizes for each dataset. We use a fixed random seed of 42 for shuffled splits.

\begin{table}[h]
\caption{Dataset filtering and split statistics. We keep image based multiple choice questions for automatic answer evaluation.}
\centering
\small
\begin{tabular}{llrrrr}
\toprule
Dataset & Source split & Filtered total & Train & Val & Test \\
\midrule
MMBench & DEV English & 1,164 & 931 & 116 & 117 \\
MMMU & Test split & 9,087 & 7,269 & 908 & 910 \\
ScienceQA & Official splits & 10,332 & 6,218 & 2,097 & 2,017 \\
\bottomrule
\end{tabular}
\label{tab:dataset_filtering}
\end{table}

For MMBench, we use the English development set and keep filtered multiple choice samples. For MMMU, we use multiple choice samples with exactly one image, text only options, and non-empty question text. For ScienceQA, we keep image bearing multiple choice questions from the official splits.

\subsection{Attack Optimization Details}

Table~\ref{tab:attack_hparams} summarizes the main optimization settings. The trigger is parameterized by raw variables and mapped to valid pixel values through a sigmoid function. All images are resized to $448\times448$, and the border width is 16 pixels unless otherwise stated.

\begin{table}[h]
\caption{Main attack optimization hyperparameters.}
\centering
\small
\begin{tabular}{lcccccc}
\toprule
Setting & Optimizer & LR & Batch & Steps & Scheduler & Temperature \\
\midrule
Gemma4-E2B & Adam & 0.02 & 64 & 2000 & cosine & 3.0 \\
Qwen3-VL-2B on MMBench & Adam & 0.02 & 32 & 600 & cosine & 4.0 \\
Qwen3-VL-2B on MMMU/ScienceQA & Adam & 0.02 & 32 & 600 & cosine & 3.0 \\
\bottomrule
\end{tabular}
\label{tab:attack_hparams}
\end{table}

For ablation studies, we use Gemma4-E2B on MMBench as the base setting. We sweep temperature over $\{1,2,4,5\}$ with $T=3$ as the default, and border width over $\{4,8,32,48\}$ with width 16 as the default. We also compare the default border trigger with temperature distribution matching against a global adversarial noise variant and a border trigger optimized with a maximum entropy objective.

\subsection{Uncertainty Metric Computation}

\paragraph{Answer Token Probability.}
We use the probability assigned to the first generated token, which corresponds to the answer option under our prompt format:
\[
\mathrm{ATP} = p(y_1 \mid x_I, x_Q).
\]
Lower ATP indicates lower confidence.

\paragraph{PPL Confidence.}
For the full generated response, including the answer option and explanation, we compute the average token log probability:
\[
\mathrm{PPLConf}
=
\exp\left(\frac{1}{N}\sum_{i=1}^{N}\log p(y_i\mid y_{<i},x_I,x_Q)\right).
\]
This is equivalent to the inverse of perplexity. Lower PPL confidence indicates lower confidence.

\paragraph{Jaccard.}
For each input, we sample $m=5$ responses using temperature 1.0, top-p 0.9, and at most 64 new tokens. We split each response by whitespace into a token set $S_i$ and compute the mean pairwise Jaccard similarity:
\[
\mathrm{Jaccard}
=
\frac{1}{m^2}
\sum_{i=1}^{m}
\sum_{j=1}^{m}
\frac{|S_i\cap S_j|}{|S_i\cup S_j|}.
\]
We include self pairs in the average. Lower Jaccard indicates less consistent sampled outputs and therefore lower confidence.

\subsection{Baselines and Preprocessing Defenses}

\paragraph{Common perturbation baseline.}
The defocus blur baseline applies a disk shaped averaging kernel of radius 6 pixels followed by a Gaussian anti aliasing filter with $\sigma=0.5$ to all input images.

\paragraph{Prompt injection baseline.}
For the prompt injection baseline, we prepend the instruction ``Before answering, compare the two most plausible options and try to refute your top choice with one plausible counterargument. Then commit to the final answer.'' to the textual query.

\paragraph{Preprocessing defenses.}
For JPEG compression, we use quality 75. For additive Gaussian noise, we use $\sigma=0.04$ on $[0,1]$ normalized pixels. For bit depth reduction, we quantize 8 bit images to 5 bit color depth.

\section{Additional Results}
\label{app:additional_results}

This section provides additional results that complement the main paper. We include results for other uncertainty metrics, model families, and datasets that follow the same evaluation protocol as in Section~\ref{sec:main_results}. Overall, these results show trends consistent with the main findings: FDA increases strong-model deferral while largely preserving final answer quality.

\begin{figure}[!t]
  \centering
  \includegraphics[
    width=\linewidth,
    trim=0 10 0 0,
    clip
  ]{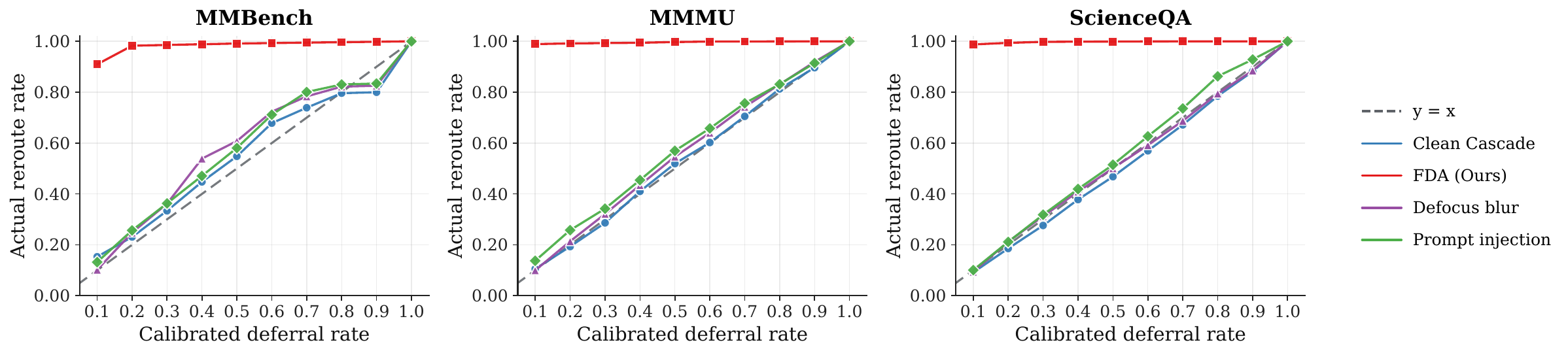}
 \caption{Actual reroute rate of the Gemma same family cascade under AP.}
\end{figure}

\begin{figure}[!t]
  \centering
  \includegraphics[
    width=\linewidth,
    trim=0 10 0 0,
    clip
  ]{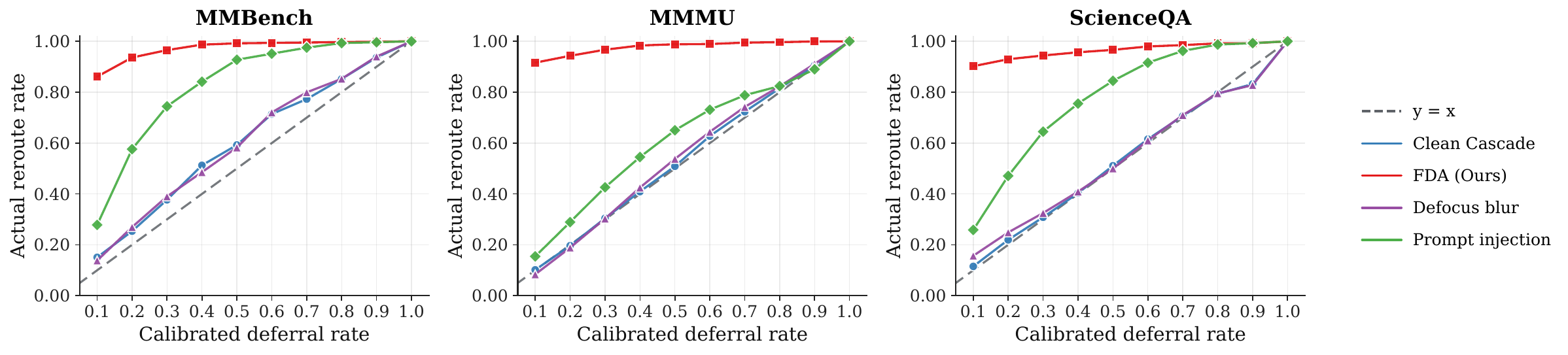}
 \caption{Actual reroute rate of the Gemma same family cascade under JS.}
\end{figure}

\begin{figure}[!t]
  \centering
  \includegraphics[
    width=\linewidth,
    trim=0 10 0 0,
    clip
  ]{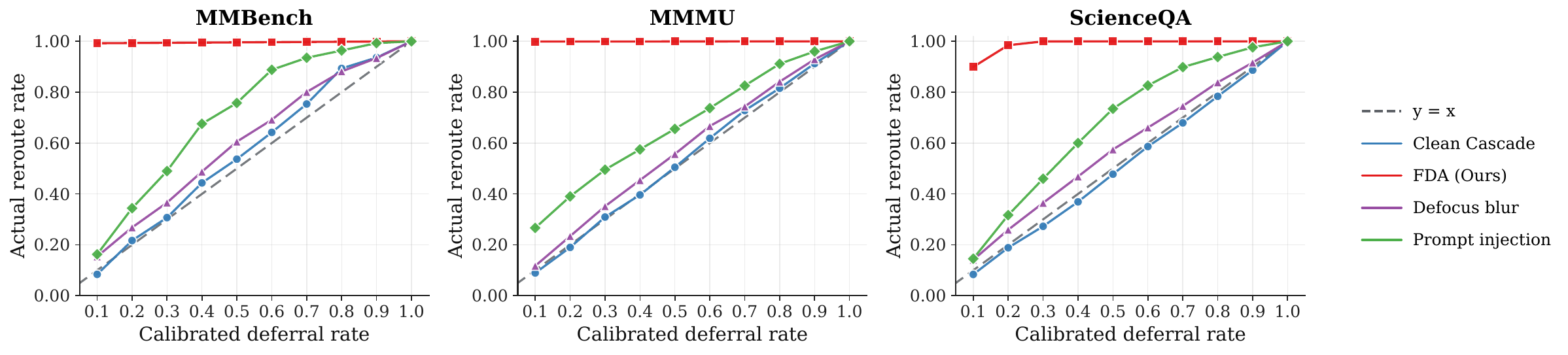}
 \caption{Actual reroute rate of the Qwen3-VL same family cascade under AP.}
\end{figure}

\begin{figure}[!t]
  \centering
  \includegraphics[
    width=\linewidth,
    trim=0 10 0 0,
    clip
  ]{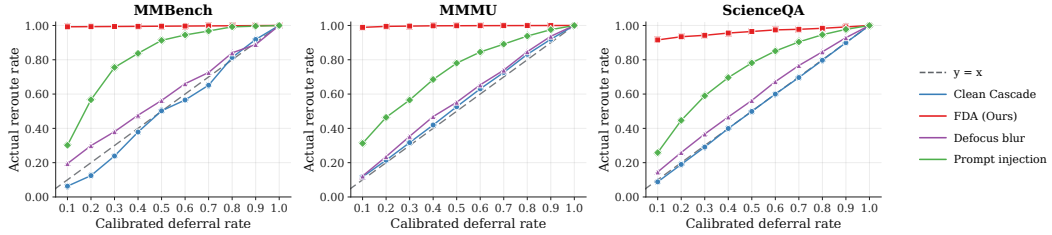}
 \caption{Actual reroute rate of the Qwen3-VL same family cascade under JS.}
\end{figure}

\begin{figure}[ht]
  \centering
  \includegraphics[
    width=\linewidth,
    trim=0 10 0 0,
    clip
  ]{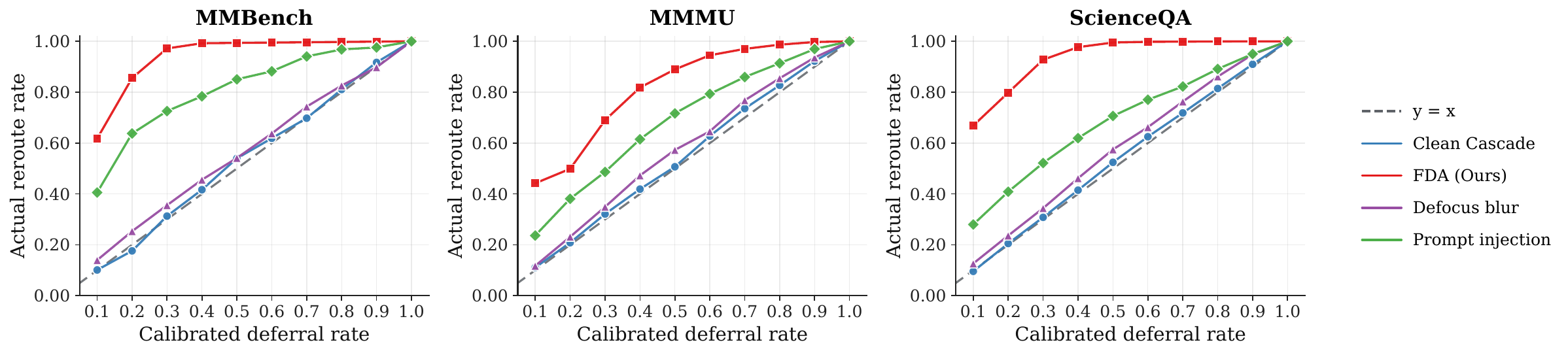}
 \caption{Actual reroute rate of the Qwen3-VL same family cascade under 1/PPL.}
\end{figure}

\end{document}